\pdfoutput=1
\documentclass[runningheads]{llncs}
\usepackage[T1]{fontenc}
\usepackage{graphicx}
\usepackage{amsfonts, amsmath}
\usepackage{algorithm}
\usepackage{algpseudocode}
\usepackage{cite}
\usepackage{subcaption}
\usepackage{booktabs}
\usepackage{color}
\usepackage[colorlinks=true,
            linkcolor=red,
            citecolor=blue]{hyperref}
\begin{document}
\title{\texttt{SeLoRA}: Self-Expanding Low-Rank Adaptation of Latent Diffusion Model for Medical Image Synthesis}
\titlerunning{Self Expanding Low-Rank Adaptation of LDM for Medical Image Synthesis}

\author{Yuchen Mao\inst{1} \and
Hongwei Li\inst{2} \and
Wei Pang\inst{3} \and
Giorgos Papanastasiou\inst{4} \and
Guang Yang\inst{5} \and
Chengjia Wang\inst{3}
}% %

\institute{University of Edinburgh
\and Harvard Medical School
\and Heriot-Watt University
\and Archimedes Unit, Athena Research Centre
\and Imperial College London}
\authorrunning{Mao et al.}
% \author{Yuchen Mao}
% %
% \authorrunning{Anonymous}
% \institute{University of Edinburgh}
% \institute{Harvard Medical School}
% \email{***@***.***}}
\maketitle              % typeset the header of the contribution
\begin{abstract}
% The persistent challenge of medical image synthesis posed by the scarcity of annotated data underscored the imperative development of effective the synthesis methods. Existing models often tarined from scratch for a single modality, making seamless transitions between different domains a formidable task. In this paper, we introduce a novel Low rank adaptation (LoRA) method - the Self-Expanding Low Rank Adaptation Module - to harnessing the power of latent diffusion models (LDMs) pre-trained on natural images, thereby facilitating the synthesis of scarce medical images. This module, designed to seamlessly integrate with the intricate LDM architecture, allows LoRA to autonomously grow to different rank across layers, mitigating LoRA's challenges associated with uniform rank selection. The proposed method not only adapts efficiently to medical images but also empowers the model to achieve optimal performance with minimal rank. The inherent flexibility of LoRA further enables efficient interchange for image synthesis across various medical domains.

The persistent challenge of medical image synthesis posed by the scarcity of annotated data and the need to synthesize `missing modalities' for multi-modal analysis, underscored the imperative development of effective synthesis methods. Recently, the combination of Low-Rank Adaptation (LoRA) with latent diffusion models (LDMs) has emerged as a viable approach for efficiently adapting pre-trained large language models, in the medical field. However, the direct application of LoRA assumes uniform ranking across all linear layers, overlooking the significance of different weight matrices, and leading to sub-optimal outcomes. Prior works on \emph{LoRA} prioritize the reduction of trainable parameters, and there exists an opportunity to further tailor this adaptation process to the intricate demands of medical image synthesis. In response, we present \emph{SeLoRA}, a Self-Expanding Low-Rank Adaptation Module, that dynamically expands its ranking across layers during training, strategically placing additional ranks on crucial layers, to allow the model to elevate synthesis quality where it matters most. The proposed method not only enables LDMs to fine-tune on medical data efficiently but also empowers the model to achieve improved image quality with minimal ranking. The code of our \emph{SeLoRA} method is publicly available on \url{https://anonymous.4open.science/r/SeLoRA-980D}.

\end{abstract}
\section{Introduction}

% The scarcity of high-quality medical image datasets, attributable to privacy concerns and the need of trained medical experts for annotation, has persisted as a significant issue. 

Foundation models \cite{tu2024towards,huang2023visual,zhou2023foundation} are increasingly gaining traction in medical imaging, offering a new paradigm for data processing and analysis. While most foundational models are trained with large natural-image datasets \cite{tu2024towards}, such as \emph{ImageNet}, the shortage of medical images is increasingly problematic \cite{5206848}. Medical image synthesis presents a valid approach to address this issue by generating synthetic images to expand and enhance scarce image datasets. 
% Beyond dataset expansion, medical image synthesis can also anonymize datasets, increase data diversity, and create photo-realistic representations for exploring and studying rare cases.
Challenges in medical image synthesis have been widely explored, leading to various proposed methods. These models are typically trained from scratch for a single modality like brain MRI \cite{Dalmaz_2022}, lung CTs\cite{MENDES2023119350}, Cataract Surgery Samples \cite{frisch2023synthesising}, and others. % can be chanegd to other work. 
However, training models from scratch requires long training times and can lead to performance limitations due to dataset size.

Recognizing the success of pre-trained foundational models on natural images, recent works have shifted towards adapting these models, such as stable diffusion \cite{Rombach_2022_CVPR}, to enable more efficient training for medical image synthesis. For example, leveraging text-based radiology reports as a condition to incorporate detailed medical information, and fine-tuning latent diffusion models has achieved significant performance gains \cite{chambon2022roentgen, chambon2022adapting}. Further exploration involves the incorporation of parameter-efficient fine-tuning (PEFT) methods, which not only makes fine-tuning more efficient but also demonstrates superior performance compared to full fine-tuning \cite{dutt2023parameterefficient}. In this context, we place a particular emphasis on applying the Low-rank Adaptation (LoRA) \cite{hu2022lora} method, a type of PEFT method, for fine-tuning stable diffusion in medical image synthesis.

Originally designed for adapting large language models, \emph{LoRA} hypothesizes that the weight matrix updated during fine-tuning exhibits a low `intrinsic rank'. Therefore, it prompted to use the product of two trainable low-rank decomposition matrices to mimic the weight update for a specific task, expressed as follows:
\begin{equation} \label{eq:lora}
    W = W_0 + AB,
\end{equation}
where the weight matrix, $W^{d_{in} \times d_{out}}$, is updated by combining the frozen original weight $W_0^{d_{in} \times d_{out}}$, and the product of two trainable low-rank decomposition matrices, $A \in \mathbb{R}^{d_{in} \times r}$ and $B \in \mathbb{R}^{r \times d_{out}}$. The rank, $r$, is deliberately chosen to be significantly smaller than the dimension of the original weight. Consequently, the trainable parameters under this configuration constitute a fraction of $W_0$'s parameter count, achieving parameter-efficient fine-tuning.

The design of \emph{LoRA} introduced the challenges of selecting the optimal rank. Small ranks yield suboptimal performance, while large ranks escalate parameter count, and searching for optimal rank for each individual \emph{LoRA} on different layers is computationally expensive. In response, \emph{LoRA} proposes to select a uniform rank across layers, but this intensifies the problem associated with rank when \emph{LoRA} is applied to LDMs. The bottleneck nature of denoising \emph{U-Net} implies a diversity in the shape of weight matrices. Thereby, the underlying rank of the weight update may also vary. Consequently, uniform rank selection on LDMs may introduce exacerbated results, ultimately compromising the quality of synthesized images.

In addressing the challenge of rank selection and aiming to achieve superior synthesized image quality with minimal trainable parameters, we draw inspiration from the concept of self-expanding a neural network \cite{mitchell2024selfexpanding}. Departing from approaches of setting a predefined uniform rank, our approach advocates for dynamically expanding the rank of \emph{LoRA} to better align with the unique needs of each layer. As a result, in our work, we present a Self-Expanding Low-rank Adaptor (SeLoRA), akin to \emph{LoRA}'s structure but distinguished by the dynamic growth of ranks guided by Fisher information during training. This enables \emph{SeLoRA} to flexibly adapt to the inherent characteristics of each layer, guaranteeing enhanced medical image synthesis quality while minimizing challenges related to rank adjustments.

\section{Related Work on {LoRA}}
This section focuses on reviewing low-rank adaptation methods that address the rank selection problem.
Building upon \emph{LoRA}, several lines of research are being developed. While many approaches primarily aim to reduce the number of trainable parameters using techniques like quantization \cite{dettmers2023qlora}, shared weights \cite{kopiczko2024vera}, and various product operations \cite{yeh2024navigating}, there is limited work on how to enhance the model performance or tackle the rank selection problem as primary objectives. 

\begin{figure}[!t]
    \centering
    \includegraphics[width=\textwidth]{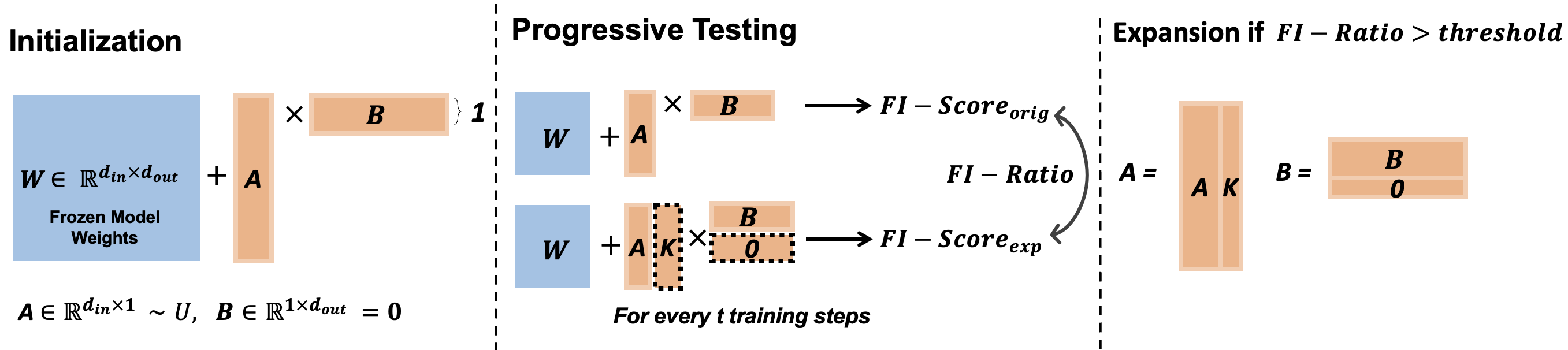}
    \caption{Training illustration of a single \emph{SeLoRA}. \emph{SeLoRA} behaves similarly to a basic \emph{LoRA} during training. However, it is tested for the expanded rank every $t$ step and is progressively expanded if the FI-Ratio exceeds the desired threshold.}
    \label{fig:training-illustiation}
\end{figure}

Among these works, one line of research explores adaptive ranking to address selection issues, wherein each layer learns to have a different rank through pruning. \emph{DyLoRA} utilizes nested dropout loss to sort the rank, allowing \emph{LoRA} to be pruned to the optimal rank after training \cite{Valipour2022DyLoRAPT}. \emph{AdaLoRA} assigns an importance score for each rank, evaluated based on the magnitude of singular values, thereby pruning less important ranks \cite{zhang2023adaptive}. \emph{SoRA} introduces a specially designed gating function that zeros out useless rank during training \cite{ding2023sparse}. While all these pruning-based methods effectively achieve adaptive rank across layers, the fundamental challenge of rank selection remains unsolved, as they introduce a fixed upper bound, which is the initial starting rank of the trainable low-rank matrices.

\section{SeLoRA}
% \subsection{Problem Formulation}
% In addressing the challenge of adapting stable diffusion for efficient high-quality medical image synthesis, our primary goal is to develop an improved LoRA. This enhanced LoRA aims to dynamically expand its rank during training to accommodate the unique needs of each layer. By doing so, we intend to enhance overall performance while keeping minimum trainable parameter.

In addressing the challenge of rank selection and maintaining superior performance with minimum rank, we propose \emph{Self-Expanding Low-Rank Adaptation} (SeLoRA). The general idea of \emph{SeLoRA} is to initialize the trainable low-rank decomposition matrices with rank $r = 1$ and dynamically expand its rank individually during training to adapt to varying layer needs. Now, to facilitate adaptive growth, we address two key research questions: (1) How can to expand the rank \textbf{ without perturbing the output}? (2) At what juncture should the expansion occur?

\subsubsection{How to expand?}
When \emph{SeLoRA} expands itself by adding a new rank, the model's final prediction should remain constant. Hence, to prevent perturbations in the model's output when introducing a new rank, a straightforward approach is to force the product of the expanded rank to be $0$. However, simple all-zero initialization poses a challenge as it hinders gradient flow through the expanded rank during initial back-propagation. Consequently, utilizing methods e.g., Fisher information to assess the expansion is not applicable. Therefore, we propose to initialize the expanded rank of matrix $B$ to be $0$, and the expanded sub-matrix, $K$, of matrix $A$ to be initialized with Kaiming uniform initialization. Hence, the expanded form of \emph{SeLoRA} can be expressed as follows:

\begin{equation} \label{eq:update}
    f(x) = x W_0 + x
    \begin{bmatrix} A & K \end{bmatrix}	
    \begin{bmatrix} B \\ 0\end{bmatrix}	
+ b_0 
\end{equation}
where $K \in \mathbb{R}^{d_{in} \times 1}$ is a vector initialized with Kaiming uniform initialization. Now, the expanded \emph{SeLoRA} maintains the desired output while allowing the gradient to propagate through $A$.

\subsubsection{When to expand?}
To determine when an expansion could enhance the model, a crucial criterion is evaluating the potential improvement introduced by the rank addition. In assessing the viability of an expanded \emph{SeLoRA} without excessive training, we employ Fisher information to measure the information conveyed by \emph{SeLoRA} weights from the datasets. The conceptualisation of using Fisher information for expansion decision, stems from prior works on model selection procedures (using Fisher information), such as the Akaike information criterion, the Bayesian information criterion, and later work on low-rank approximation on neural network's weights \cite{1100705, e0988035-67ff-3d00-8d22-b00bd6518fb4, hsu2022language}. Here we utilized the empirically estimated Fisher information, introduced in Equation \ref{eq:fisher-information}, as deriving an exact value is generally intractable due to the need for marginalization over the entire dataset.

\begin{equation} \label{eq:fisher-information}
\hat I_w = \frac{1}{|B|} \sum^{|B|}_{i = 1} \left(\frac{\partial}{\partial w} \mathcal{L}(b_i;w)\right)^2,
\end{equation}
where $|B|$ is the batch size, and $b_i$ is a sample in the batch. To quantify the information carried by a single \emph{SeLoRA}, we introduce the Fisher information score (FI-Score), which sums over all Fisher information of the weight matrices $A, B$ in \emph{SeLoRA}, as shown in Equation \ref{eq:fisher-score}:

\begin{equation}
    \label{eq:fisher-score}
    \text{FI-Score} = \sum_{i = 1}^{d_{in}} \sum_{j = 1}^{r} \hat I_{A_{i,j}}  +\sum_{i = 1}^{r} \sum_{j = 1}^{d_{out}} \hat I_{B_{i,j}}.
\end{equation}

Here, while the FI-Score is calculated by back-propagating gradients, it is important to note that the optimizer does not update the parameters during this calculation for the expanded \emph{SeLoRA}. To further assess whether the expanded \emph{SeLoRA} is superior to the previous unexpanded version, we introduce the Fisher information Ratio (FI-Ratio). The ratio between the Fisher information score of the expanded and original \emph{SeLoRA} is calculated, as shown in Equation \ref{eq:fisher-ratio}:

\begin{equation} \label{eq:fisher-ratio}
    \text{FI-Ratio} = \frac{\text{FI-Score}_{orig}}{\text{FI-Score}_{exp}}.
\end{equation}

Equation \ref{eq:fisher-ratio} provides a metric to evaluate the information gained when \emph{SeLoRA} is expanded. This ratio is measured at each \emph{SeLoRA} module, and \emph{SeLoRA} undergoes expansion when the FI-Ratio exceeds a desired threshold $\lambda$. An aggressive threshold, such as $\lambda = 1$, accepts any improvement to the model. Alternatively, a conservative threshold, such as $\lambda = 1.3$ or a larger value, could result in a more cautious expansion. Additionally, to ensure each expansion is beneficial, a hyperparameter $t$ is introduced. Testing of \emph{SeLoRA}'s expansion is conducted at each $t$ training step, allowing the previous expanded rank to learn and converge before testing the new expansion. The detailed training procedure is shown in Algorithm \ref{alg:SeLoRA}.

\begin{algorithm}[!t]
\caption{\emph{SeLoRA} Training Procedure}\label{alg:SeLoRA}
\begin{algorithmic}[1]

\State \textbf{Initialization:} Initialize $W = W_0 + AB$ for each linear layer with $A$, $B$ at rank $r=1$.

\State \textbf{Training:}
\While{$s < \text{Total Steps}$}
    \State Forward pass $h = x(W_0 + AB) + b_0$.
    \State Update $A$ and $B$ with gradients $\nabla_A \mathcal{L}$ and $\nabla_B \mathcal{L}$.
    \If{$s \mod t = 0$} 
        \For{each \emph{SeLoRA} module}
            \State Compute original and expanded FI-Scores for $A$, $B$ and $A'$, $B'$.
            \State $A' = \begin{bmatrix} A & K \end{bmatrix}$, $B' = \begin{bmatrix} B^T & 0\end{bmatrix}^T$.
            \State FI-Ratio $= \frac{\text{FI-Score}_{\text{orig}}}{\text{FI-Score}_{\text{exp}}}$.
            \If {FI-Ratio $\geq \lambda$} Update $A \leftarrow A'$, $B \leftarrow B'$.
            \EndIf
        \EndFor
    \EndIf
    \State $s \leftarrow s + 1$.
\EndWhile

\end{algorithmic}
\end{algorithm}

\subsection{Datasets for Evaluation}

We compare \emph{SeLoRA} against \emph{LoRA} and other baselines, assessed on two metrics. A rank allocation analysis is then performed to examine the distribution of ranks within the trained \emph{SeLoRA} model, followed by an ablation on different $\lambda$ values.
% \subsection{Datasets and Experiments setup}
% For our experiments, we've used IU X-RAY \cite{DemnerFushman2015PreparingAC} and Montgomery County CXR datasets \cite{data7070095}.

\vspace{0.3cm}
\noindent
\textbf{IU X-RAY Dataset} \cite{data7070095}, collected by Indiana University, consists of 3,955 radiology reports paired with Frontal and Lateral chest X-ray images, each accompanied by text findings. For our experiments, we arbitrarily chose to exclusively focus on frontal projections, considering the substantial dissimilarities between frontal and lateral images. The dataset is partitioned into training, validation, and testing sets, with $80\%/16\%/4\%$ of the datasets, respectively.

\vspace{0.3cm}
\noindent
\textbf{Montgomery County CXR Dataset} is relatively compact, consisting of 138 frontal chest X-rays \cite{DemnerFushman2015PreparingAC}. Each entry in the dataset includes an X-ray image, along with details about the patient's sex, age, and medical findings. Among the 138 cases, 80 are labeled as `normal,' with the remaining cases describing specific patient illnesses. The patient's sex, age, and findings are concatenated as prompt. The dataset is then divided into training and testing sets with a split ratio of $80\%/20\%$. Omitting a validation set ensures a sufficient number of samples for both the training and testing phases.
\\

During fine-tuning, to accommodate the text encoder of the stable diffusion, any text finding longer than 76 tokens is truncated, as outlined in the \emph{CLIP} approach \cite{DBLP:journals/corr/abs-2103-00020}. As for the training images, they are initially resized such that the shortest side is 224 pixels wide, followed by centered cropping to obtain an image size of $224 \times 224$.

\subsection{Implementation Details}
The experiments are implemented in Pytorch, utilizing the base stable diffusion model weights obtained from `runwayml/stable-diffusion-v1-5' in huggingface~\cite{Rombach_2022_CVPR}. \emph{LoRA}, \emph{DyLoRA}, \emph{AdaLoRA}, and \emph{SeLoRA} are used in experiments for comparison. All \emph{LoRA} modules are only injected into the linear layer of the stable diffusion, which consists of the \emph{U-Net} and the text encoder used to condition the denoising process. Prior works have proved that fine-tuning the VAE part of the stable diffusion model leads to no significant differences in performance on medical images, so injecting the \emph{LoRA} module into VAE was omitted \cite{chambon2022adapting}. The models are trained using mean squared error loss and are evaluated after each epoch on the validation set. The best-performing model is then utilized to examine inference on the test set. Model trains for 10 epochs on IU-Xray dataset and 100 epochs on Montgomery County CXR due to its smaller size. For our method, we chose to use a threshold of $\lambda = 1.1$, accompanied by $t = 40$ training steps. While these parameter choices are arbitrary, a relatively large $t$ step is desired to allow \emph{SeLoRA} to fully converge before the evaluation of expansion. For \emph{LoRA} and \emph{DyLoRA}, the rank is set to $r = 4$, and the rank of \emph{AdaLorA} is initialised with rank $r = 6$ and reduced to the target rank of $r = 4$. The original \emph{LoRA} experiments was conducted with a rank of $r = \{1, 2, 4, 8, 64\}$. We selected the median value, $r = 4$ across all testing procedures, as it has a comparable number of trainable parameters to our trained \emph{SeLoRA}, ensuring a fair comparison. 

\subsection{Evaluation Metric}
We assessed the synthetic images based on both the fidelity and their alignment with the provided prompts. To measure fidelity, we use Fréchet Inception Distance (FID), which calculates the similarity between the distribution of synthetic datasets and the distribution of the original datasets \cite{10.5555/3295222.3295408}. 
% Additionally, Structural Similarity Index (SSIM) is used, which measures the quality based on the computation of three factors: luminance, contrast, and structure \cite{1284395}. 
For assessing alignment with the prompt, CLIP score is used, which measures text-to-image similarity \cite{DBLP:journals/corr/abs-2103-00020}. Results are averaged across three random seeds and summarized in Table \ref{tab:merged} for IU X-RAY and Montgomery County CXR Datasets, respectively.

\section{Results}

% \begin{table}[!t]
% \caption{Performance Evaluation for IU X-RAY Dataset. The best result is indicated in bold.}
% \label{tab1} \centering
% \begin{tabular}{l|l l l}
% \hline
% Methods  \quad &\quad FID $\downarrow$  \quad &\quad CLIP Score $\uparrow$\\
% \hline
% LoRA  \quad &\quad$113.365 \pm 7.881 $\quad &\quad $27.112 \pm 0.177$\\ 
% AdaLoRA \quad &\quad $184.248 \pm 53.046$ \quad &\quad $26.895 \pm 0.329$  \\ 
% DyLoRA \quad &\quad $116.032 \pm 10.289$  \quad &\quad $26.885 \pm 0.049$  \\ 
% SeLoRA \quad &\quad \textbf{113.042 $\pm$ 16.794}\quad &\quad \textbf{27.256$ \pm $0.222}\\ 
% \hline
% \end{tabular}
% \end{table}

% \begin{table}[!t]
% \caption{Performance Evaluation for Montgomery County CXR Dataset. The best result is indicated in bold.}
% \label{tab2} 
% \centering
% \begin{tabular}{l|l l l}
% \hline
% Methods  \quad &\quad FID $\downarrow$ \quad &\quad CLIP Score  $\uparrow$ \\
% \hline
% LoRA \quad &\quad $461.795 \pm 14.099$ \quad &\quad $24.336 \pm 0.708$ \\ 
% AdaLoRA \quad &\quad $489.238 \pm 14.277$  \quad &\quad $22.473 \pm 0.233$ \\ 
% DyLoRA \quad &\quad $475.423 \pm 7.844$  \quad &\quad $ 22.932 \pm 0.267$ \\ 
% SeLoRA  \quad &\quad \textbf{205.537 $\pm$ 9.555}  \quad &\quad \textbf{26.380 $\pm$ 0.041}\\ 
% \hline
% \end{tabular}
% \end{table}

\begin{table}[!t]
\caption{Quantitative results of IU X-RAY and CXR datasets. The best results are highlighted in bold.}
\label{tab:merged} 
\centering
\begin{tabular}{l | l l | l l }
\hline
 & \multicolumn{2}{c}{IU X-RAY} & \multicolumn{2}{c}{Montgomery County CXR}\\
\hline
Methods  \quad &\quad FID $\downarrow$ \quad &\quad CLIP Score  $\uparrow$ \quad &\quad FID $\downarrow$ \quad &\quad CLIP Score  $\uparrow$ \\
\hline
LoRA  \quad &\quad$113.37 \pm 7.88 $\quad &\quad $27.11 \pm 0.18$      \quad &\quad $461.80 \pm 14.10$ \quad &\quad $24.34 \pm 0.71$  \\ 
AdaLoRA \quad &\quad $184.25 \pm 53.05$ \quad &\quad $26.90 \pm 0.33$  \quad &\quad $489.24 \pm 14.28$  \quad &\quad $22.47 \pm 0.23$ \\ 
DyLoRA \quad &\quad $116.03 \pm 10.29$  \quad &\quad $26.89 \pm 0.05$  \quad &\quad $475.42 \pm 7.84$  \quad &\quad $ 22.93 \pm 0.27$ \\ 
SeLoRA \quad &\quad \textbf{113.04 $\pm$ 16.79}\quad &\quad \textbf{27.26$ \pm $0.22}  \quad &\quad \textbf{205.54 $\pm$ 9.56}  \quad &\quad \textbf{26.38 $\pm$ 0.04}\\ 
\hline
\end{tabular}
\end{table}

Our results clearly highlight the superior performance of the proposed method, \emph{SeLoRA}, in terms of fidelity and alignment with the prompt (text condition). In Table \ref{tab:merged}, we observe that \emph{SeLoRA} consistently outperformed all other methods across all evaluation metrics for these relatively small datasets. These results demonstrate that \emph{SeLoRA} is advantageous and truly promising to be examined and expanded across more medical image synthesis data and problems, given the inherent scarcity of medical datasets \cite{Papanastasiou_2023}.
% (aka "data challenge" in medical imaging). 
Moreover, \emph{SeLoRA} outperforms all other \emph{LoRA} variants on IU-XRay by only using small fractions of trainable parameters, averaging $0.121$ and $0.368$ for the text encoder and U-net, respectively. In contrast, other \emph{LoRA} methods with rank $r = 4$ use nearly double the trainable parameters ($0.216$ and $0.803$) while achieving lower scores. 
% It is important to note that the relatively high FID score presented in Table \ref{tab1}, compared to state-of-the-art image synthesis on natural images, is attributed to the small testing sample size. This limitation is an inherent result of the scarcity of datasets in medical images.

\begin{figure}[!t]
    \centering
    \includegraphics[width=\textwidth]{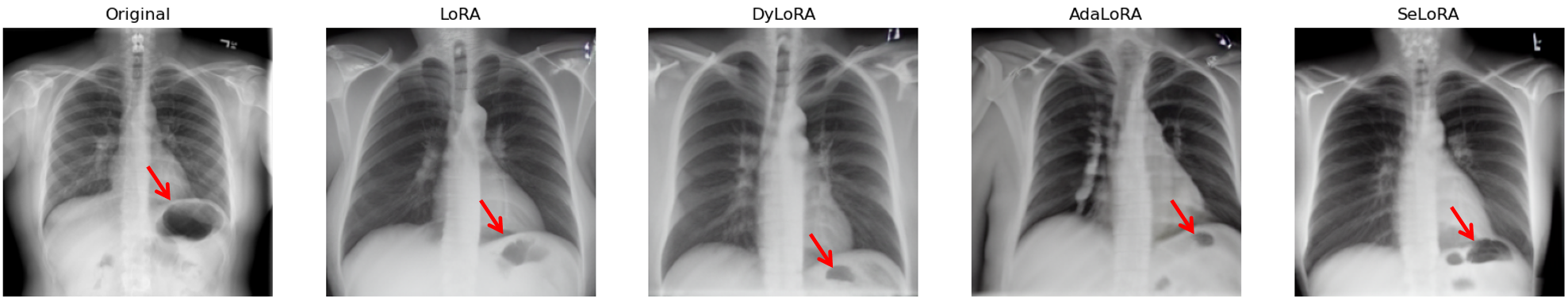}
    \caption{Qualitative comparison results obtained on the IU-Xray data, generated by fine-tuning stable diffusion models injected with various LoRA variants. Prompt used: \texttt{``Heart size and vascularity normal. These contour normal. Lungs clear. No pleural effusions or pneumothoraces.''} More sample results are presented in the Appendix.}
    \label{fig:qualitative}
\end{figure}

Furthermore, synthetic images generated by a model trained on the IU-Xray dataset are displayed in Figure \ref{fig:qualitative}. Notice that, although all synthetic images achieved similar quality, in comparison, \emph{SeLoRA} is the only method that captures the distinct pathology (represented as a black circle) in the lower part of the right lung, closely resembling the original image.

\subsection{Rank Allocation Analysis}
\label{sec:Rank_alloc}

\begin{figure}[!t]
    \centering
    \begin{subfigure}[b]{0.45\textwidth}
        \includegraphics[width=\textwidth]{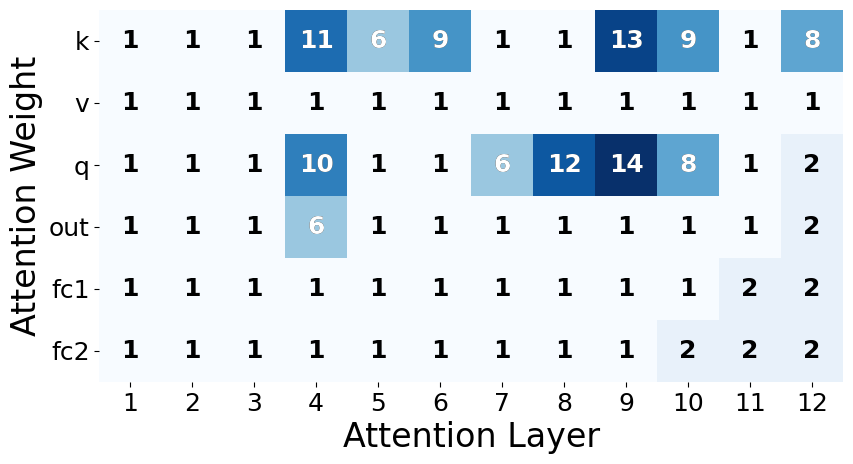}
        \caption{Rank of \emph{SeLoRA} in Text Encoder.} \label{fig1}
        \label{fig:sub1}
    \end{subfigure}
    \hfill
    \begin{subfigure}[b]{0.54\textwidth}
        \includegraphics[width=\textwidth]{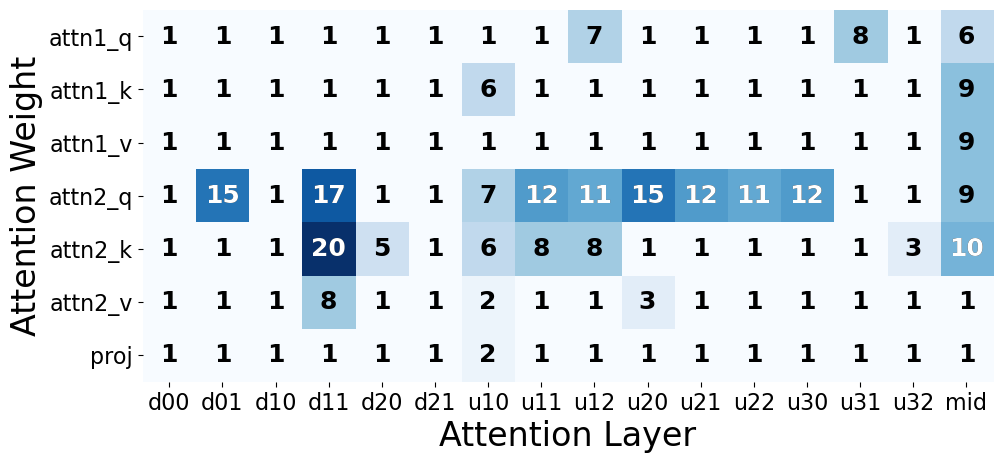}
        \caption{Rank of \emph{SeLoRA} in Denoising {U-Net}.}
        \label{fig:sub2}
    \end{subfigure}
    \caption{The final rank of \emph{SeLoRA} fine-tuned on stable diffusion with IU-XRay Dataset. The x-axis represents the layer index, and y-axis indicates the corresponding attention's weight.  \emph{SeLoRA} places more rank on crucial layers.}
    \label{fig:main}
\end{figure}

Figure \ref{fig:main} illustrates the final rank results of \emph{SeLoRA} across stable diffusion model layers after training on the IU-XRay dataset. In the text encoder part, large ranks for \emph{SeLoRA} lie at the q and k parts of the attention weights. For the U-net part, large ranks are allocated to the q and k parts of the second attention layer, namely the cross-attention layer. The rank allocation aligns with the intuition that weight updates would change most dramatically at locations where the latent representations of text and image intersect (where conditioning is more apparent). Hence, it validates our hypothesis that our proposed expansion method allows \emph{SeLoRA} to focus and place more rank on crucial layers. 

\vspace{-0.1cm}
\subsection{The Impact of $\lambda$}
\vspace{-0.1cm}
Further results obtained by varying $\lambda$ are displayed in Tab. 1, Fig. 1, and Fig. 2 in the Appendix. These results indicate that a smaller $\lambda$ leads to a more aggressive expansion of all layers in \emph{SeLoRA}, resulting in improved synthesis quality, albeit at the cost of a more expensive training process. However, the critical layers that \emph{SeLoRA} emphasizes remain consistent with our observations in Section~ \ref{sec:Rank_alloc}.

\vspace{-0.2cm}
\section{Conclusion}
\vspace{-0.15cm}
This work introduces a novel parameter-efficient method named \emph{SeLoRA}, designed to effectively fine-tune stable diffusion models for generating X-ray images based on text (radiology) prompts. Our method enables progressive expansion in the rank of \emph{LoRA}, enabling more precise image synthesis with minimal added rank. Through exploratory analysis, we demonstrate that the proposed FI-Ratio is capable of effectively guiding \emph{SeLoRA} to expand its rank and allocate more rank to crucial layers. We believe that \emph{SeLoRA}, when combined with stable diffusion, can be easily employed to adapt to various medical datasets containing text-image pairs, and potentially being applicable for clinical text synthesis. Moreover, given the increasing work on segmentation and detection in 3D (tomographic, e.g., magnetic resonance imaging, computed tomography, and other) medical imaging data and the emergence of prompt-to-3D models, in our future work, we aspire to explore \emph{SeLoRA} adaptations on fine-tuning prompt-to-3D models for 3D medical image synthesis.

% \begin{credits}
% \subsubsection{\ackname} A bold run-in heading in small font size at the end of the paper is
% used for general acknowledgments, for example: This study was funded

% \subsubsection{\discintname}
% It is now necessary to declare any competing interests or to specifically
% state that the authors have no competing interests. Please place the
% statement with a bold run-in heading in small font size beneath the
% (optional) acknowledgments\footnote{If EquinOCS, our proceedings submission
% system, is used, then the disclaimer can be provided directly in the system.},
% for example: The authors have no competing interests to declare that are
% relevant to the content of this article. Or: Author A has received research
% grants from Company W. Author B has received a speaker honorarium from
% Company X and owns stock in Company Y. Author C is a member of committee Z.
% \end{credits}
\bibliographystyle{splncs04}
\bibliography{SeLoRA}




% LNCS Homepage, \url{http://www.springer.com/lncs}, last accessed 2023/10/25
% \end{thebibliography}
\end{document}